\pdfoutput=1

\documentclass[11pt]{article}

\usepackage[preprint]{acl}
\usepackage[normalem]{ulem}
\useunder{\uline}{\ul}{}
\usepackage{tcolorbox}
\usepackage{comment}
\usepackage{times}
\usepackage{latexsym}
 \usepackage{multirow}
 \usepackage{booktabs}
 \usepackage{algorithm}
 \usepackage{algpseudocode}
 \usepackage{algorithmicx}
 \usepackage{pifont}
 \newcommand{\cmark}{\ding{51}}%
\newcommand{\xmark}{\ding{55}}%

\usepackage[T1]{fontenc}
\usepackage{enumitem}

\usepackage[utf8]{inputenc}
\usepackage{amsmath} 
\usepackage{microtype}

\usepackage{inconsolata}
\usepackage{dsfont}
\usepackage{graphicx}
\usepackage{tabularx}
\newcolumntype{Y}{>{\hsize=2\hsize}X}
\title{Joint Evaluation of Answer and Reasoning Consistency for \\Hallucination Detection in Large Reasoning Models}

\usepackage{authblk}
\author[1,2]{\textbf{Changyue Wang}\thanks{cy-wang24@mails.tsinghua.edu.cn}}
\author[1]{\textbf{Weihang Su}}
\author[2,1]{\textbf{Qingyao Ai}\thanks{Corresponding Author: aiqy@tsinghua.edu.cn}}
\author[1]{\textbf{Yiqun Liu}}

\affil[1]{Department of Computer Science and Technology, Tsinghua University}
\affil[2]{Quan Cheng Laboratory}

\begin{document}
\maketitle

\begin{abstract}
Large Reasoning Models (LRMs) extend large language models with explicit, multi-step reasoning traces to enhance transparency and performance on complex tasks.     
However, these reasoning traces can be redundant or logically inconsistent, becoming a new and hard-to-detect source of hallucination.
Existing hallucination detection methods focus primarily on answer-level uncertainty and often fail to detect hallucinations or logical inconsistencies arising from the model’s reasoning trace.
This oversight is particularly problematic for LRMs, where the explicit thinking trace is not only an important support to the model's decision-making process but also a key source of potential hallucination. 
To this end, we propose RACE (Reasoning and Answer Consistency Evaluation), a novel framework specifically tailored for hallucination detection in LRMs.
RACE operates by extracting essential reasoning steps and computing four diagnostic signals: inter-sample consistency of reasoning traces, entropy-based answer uncertainty, semantic alignment between reasoning and answers, and internal coherence of reasoning. 
The joint utilization of these signals makes RACE a more robust detector of hallucinations in LRMs.
Experiments across datasets and different LLMs demonstrate that RACE outperforms existing hallucination detection baselines, offering a robust and generalizable solution for evaluating LRMs.\footnote{The source code is available at https://github.com/bebr2/RACE}
\end{abstract}

\section{Introduction}

Large Reasoning Models (LRMs) have recently emerged as a subclass of large language models (LLMs) specifically optimized for long-sequence, stepwise reasoning~\cite{deepseekai2025deepseekr1incentivizingreasoningcapability, qwq32b, glm2024chatglm, o12024, gemini2025}. 
These models are trained with large-scale supervised fine-tuning and reinforcement learning to produce not only final answers, but also explicit and often lengthy reasoning traces that outline the model’s decision-making process. 
Such traces have demonstrated clear advantages in complex tasks such as multi-hop question answering, code generation, and mathematical problem solving~\cite{zhong2024evaluationopenaio1opportunities}.
While these explicit reasoning traces enhance transparency and often boost task performance, they also introduce new challenges: they can be redundant or logically inconsistent~\cite{cotnotfaithful, chen2025think23overthinkingo1like}.
Such issues may lead the model to produce factually incorrect or misleading conclusions, even if the reasoning appears plausible. 
Consequently, hallucination detection~\cite{zhang2023siren,manakul2023selfcheckgpt,su2024mitigating,su2024unsupervised} for LRMs must consider not only whether the final answer is correct, but also whether the underlying reasoning trace is coherent and well-grounded.

Despite these challenges, most existing black-box hallucination detection methods focus solely on output-level uncertainty, typically by sampling multiple answers and measuring their consistency~\cite{seu, sindex}.
Techniques like SelfCheckGPT~\cite{manakul2023selfcheckgpt} and Semantic Entropy~\cite{se} follow this paradigm and are effective when model outputs are short and self-contained. 
However, such methods fail when applied to LRMs, where a significant portion of the inference behavior is embedded within the reasoning trace~\cite{ke2025surveyfrontiersllmreasoning}. 
In practice, multiple samples may converge on the same final answer but follow divergent or incoherent reasoning paths. 
These reasoning inconsistencies often signal hallucinations that remain undetected when evaluation is limited to final answers.
Therefore, hallucination detection in LRMs demands a broader perspective that explicitly incorporates both the model's reasoning trace and its alignment with the final answer.

To this end, we propose \textbf{RACE} (\textbf{\underline{R}}easoning and \textbf{\underline{A}}nswer \textbf{\underline{C}}onsistency \textbf{\underline{E}}valuation), a novel black-box hallucination detection framework designed for LRM,  which explicitly integrates both the reasoning trace and the final answer into a unified evaluation. 
RACE moves beyond traditional answer-level approaches by jointly evaluating the model’s full reasoning–answer behavior. 
It decomposes hallucination detection into four complementary components: 
(1) reasoning consistency, which captures the diversity and coherence of reasoning traces across multiple generations; 
(2) answer uncertainty, measured via refined semantic entropy estimation; 
(3) reasoning–answer alignment, which evaluates whether the LRM’s main reasoning trace, when treated as context, consistently leads to the sampled final answers with high predictive confidence, indicating that the rationale genuinely supports the model-generated answer space; and
(4) reasoning internal coherence, measuring the proportion of speculative content within the reasoning path. 
To mitigate the impact of noise in reasoning paths, RACE further includes a chain-of-thought (CoT)~\cite{wei2022chain} Extraction module that distills the most relevant reasoning steps for each answer. 
This integrated evaluation enables RACE to detect hallucinations that previous methods may overlook, especially in cases where the sampled final answers appear semantically consistent but are derived from flawed reasoning.

We validate RACE's effectiveness across various datasets under black- or gray-box settings, where neither our method nor competing baselines require task-specific fine-tuning. 
We compare RACE with a wide range of existing hallucination detection methods, including probability-based metrics such as Length-Normalized Predictive Entropy (LNPE)~\cite{malinin2020uncertainty}, semantic uncertainty-based approaches (e.g., Semantic Entropy, SINdex~\cite{sindex}), and SelfCheckGPT, which focuses on sampled answer consistency. 
Across various LLMs, including both general-purpose models and those optimized for reasoning, RACE achieves the best overall performance among all baselines. 
Further ablation studies confirm the complementary value of each module in our framework and underscore the necessity of modeling reasoning traces for effective hallucination detection. 
Taken together, our results highlight RACE as a robust, generalizable solution to hallucination detection in models with explicit reasoning behavior.

To summarize, our contributions are as follows:

\begin{itemize}

\item We introduce \textbf{RACE}, a novel black-box hallucination detection framework specifically designed for Large Reasoning Models. 
Unlike prior work that focuses solely on final answers, RACE systematically incorporates both the reasoning trace and answer into a joint evaluation.

\item We design four complementary modules (reasoning consistency, answer uncertainty, reasoning–answer alignment, and reasoning internal coherence) and a reasoning distillation mechanism to jointly capture intra- and inter-sample inconsistencies for hallucination detection.

\item We conduct experiments across multiple benchmarks and model families, showing that RACE consistently outperforms existing hallucination detection baselines on LRMs, while generalizing effectively to standard LLMs.

\end{itemize}

\section{Related Works}

\begin{figure*}
    \centering
    \includegraphics[width=\textwidth]{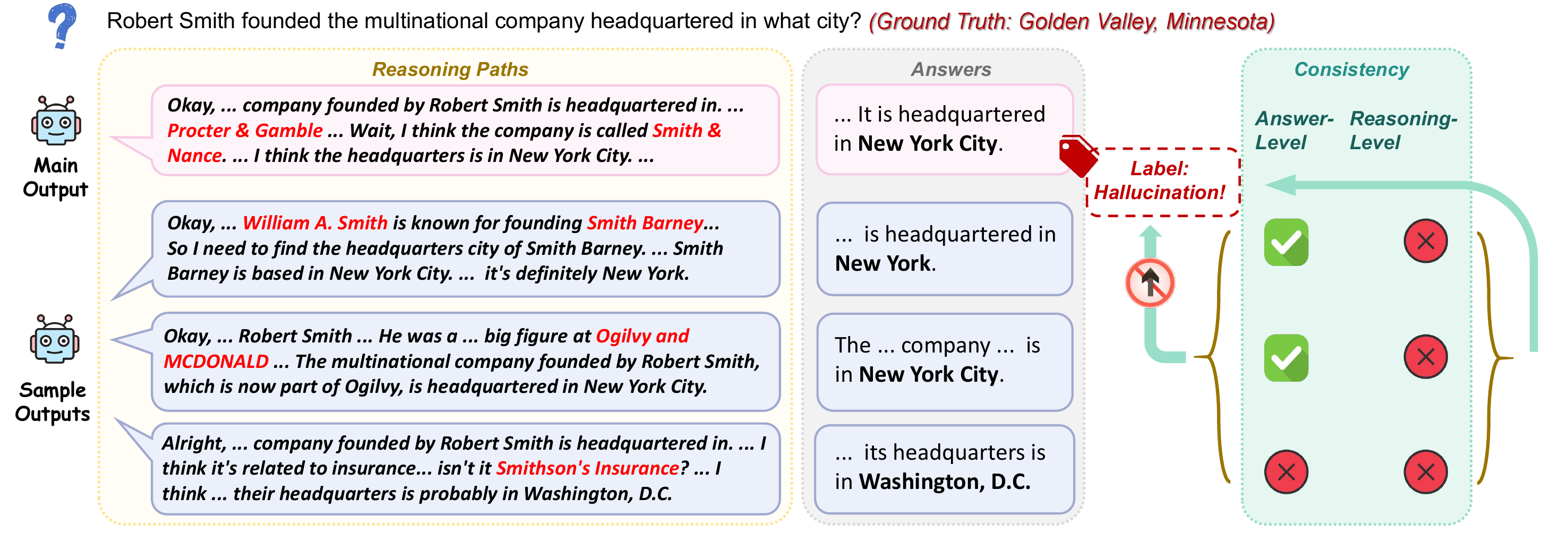}
    \caption{This figure illustrates the importance of incorporating reasoning paths in hallucination detection for LRMs. The example is from HotpotQA. The model is DeepSeek-R1-Distill-Qwen7B. While three of the four final answers mention “New York City,” the underlying reasoning traces reveal divergent and often inaccurate chains of logic. Red highlights indicate hallucinated or unsupported claims that are absent from the final answers. Thus, evaluating only the answer-level output would misleadingly suggest that the model is consistent with the answer, underscoring the need for reasoning-aware hallucination detection.}
    \label{fig:main}
\end{figure*}

\subsection{Large Reasoning Models (LRMs)}
Recent work on LRMs enhances the capabilities of LLMs by introducing explicit and multi-step reasoning mechanisms~\cite{besta2025reasoninglanguagemodelsblueprint}.
By integrating planning, reinforcement learning, and process supervision, LRMs can address more complex tasks than conventional LLMs~\cite{zhong2024evaluationopenaio1opportunities, guan2025rstarmathsmallllmsmaster}.
However, despite these gains, recent studies show that LRMs remain prone to factual hallucinations~\cite{Hughes_Vectara_Hallucination_Leaderboard_2023, lu2025auditingmetacognitivehallucinationsreasoning}, which limits their reliability in knowledge-intensive applications.

\subsection{Hallucination Detection}

This work focuses on factual hallucination, a phenomenon that LLMs can generate text that seems accurate but lacks factual accuracy ~\cite{hallusurvey}. Researchers have developed several techniques to identify these hallucinations. SAPLMA~\cite{SAPLMA} and MIND~\cite{su2024unsupervised} show that one can train a hallucination classifier based on the model's internal states, while INSIDE~\cite{INSIDE} utilizes the covariance matrix of the internal states for detection. ~\citet{ATTENTIONhallu} employs spectral properties of attention maps to identify hallucinations. However, these methods require white-box access to the LLM, limiting their widespread use. Methods based on sampling consistency are gaining traction due to their black- or gray-box nature. SelfCheckGPT~\cite{manakul2023selfcheckgpt} detects hallucinations by measuring discrepancies between sampled answers and the primary answer. Semantic Entropy ~\cite{se} uses a clustering approach, replacing token-level entropy with semantic-level uncertainty, while SINdex ~\cite{sindex} enhances this by assessing intra- and inter-cluster consistency. ~\citet{seu} propose evaluating consistency through pairwise similarity of full response embeddings. While these methods primarily target shorter outputs, they may overlook the reasoning process, leaving room to improve hallucination detection for LRMs.




\section{Foundations of the RACE Framework}
\label{sec:foundation}

In this section, we introduce an information-theoretic formulation of hallucination detection that motivates the design of our proposed \textsc{RACE} framework. While existing sampling-based approaches typically assess output uncertainty by sampling multiple answers and measuring their semantic consistency~\cite{se, manakul2023selfcheckgpt}, such methods are primarily effective for short, self-contained responses that do not involve complex intermediate reasoning.
This assumption, however, breaks down in the context of large reasoning models (LRMs), which produce explicit and often lengthy reasoning traces as part of their outputs. In these cases, a substantial portion of the model’s inference is embedded within the reasoning process itself. As a result, even when final answers appear consistent across samples, their corresponding reasoning paths can diverge significantly, revealing hallucinations that remain undetected by answer-only metrics. Figure~\ref{fig:main} demonstrates how agreement in final answers can fail to reveal inconsistencies in the underlying reasoning traces. Although most sampled outputs reach the same answer (“New York”), their reasoning paths differ substantially and contain hallucinated or unsupported claims, emphasizing the need for reasoning-aware hallucination detection.

To better characterize this phenomenon, we adopt an information-theoretic perspective, modeling the joint uncertainty over reasoning \(R\) and answer \(A\) given the question \(Q\). Specifically, we decompose the joint entropy as:

\begin{small}

\begin{equation}
\label{eq1}
H(R, A \mid Q) = H(R \mid Q) + H(A \mid Q) - I(R, A \mid Q),
\end{equation}
\end{small}

\noindent This decomposition reveals three complementary sources of uncertainty and alignment:
\begin{itemize}
    \item {\small \( H(R \mid Q) \)}: uncertainty in reasoning traces;
    \item {\small \( H(A \mid Q) \)}: variability in final predictions;
    \item {\small \( I(R, A \mid Q) \)}: consistency between reasoning and answer space, i.e., how strongly the reasoning supports the final answer space.
\end{itemize}

\noindent Based on the above formulation, it becomes natural to consider hallucination detection as a unified assessment of three factors: the variability of reasoning traces, the uncertainty of final answers, and the degree of alignment between the two. This perspective provides a principled foundation for developing more comprehensive frameworks for hallucination detection. In the next section, we present \textsc{RACE}, a practical implementation of this formulation that quantifies each component through corresponding scoring mechanisms.

\section{Methodology}

Building on the information-theoretic formulation introduced above, we now present \textsc{RACE}, a hallucination detection framework that evaluates consistency across reasoning and answer components.
In this section, we begin by defining notations, then describe a CoT Extraction module to distill key reasoning steps. Finally, we present the scoring framework that evaluates hallucination risk based on reasoning consistency, answer uncertainty, reasoning-answer alignment, and internal coherence, inspired by Equation~\ref{eq1}.

\subsection{Problem Formulation}

\label{sec:settings}

Following the paradigm of existing consistency-based hallucination detection methods, we generate a main output \(O_{\text{main}}\) using greedy decoding, which better reflects the model’s default inference-time behavior, and sample \(N\) additional outputs \(O_1, \dots, O_N\) for consistency evaluation. 
Each output \(O\) comprises a reasoning component \((R)\) and a final answer \((A)\). 
The goal is to perform a binary classification to determine whether the main answer $A_{main}$ contains factual hallucinations.
Note that RACE is designed to accommodate various model types, including not only LRMs but also standard instruction-tuned LLMs, each of which produce outputs in different formats. We consider three representative settings:

\begin{itemize}
    \item \textbf{LRMs:} The model explicitly outputs a reasoning trace \(R\) (often enclosed in tags like \texttt{<think>...</think>}).
    \item \textbf{Chain-of-Thought from Standard LLMs:} The model is prompted (e.g., “think step by step”) to generate the CoT \(R\) before concluding with the answer \(A\).
    \item \textbf{Direct Answers from Standard LLMs:} Without explicit reasoning prompts, we treat the entire output of the model concurrently as both $R$ and $A$, i.e., $R=A=O$.
\end{itemize}

After identifying $R$ and $A$, we apply a CoT Extraction module to distill key reasoning steps and reduce noise. 
This serves as a foundation for the scoring framework introduced in Equation~\ref{eq1} and is detailed in the next subsection.

\subsection{Reasoning Trace Distillation}

Reasoning traces generated by LRMs often include exploratory thoughts or redundant steps that do not directly contribute to the final answer~\cite{chen2025think23overthinkingo1like}. 
Such reasoning noise can obscure consistency assessment and reduce the reliability of hallucination detection.
To address this, we propose a specialized \textbf{CoT Extraction} module designed to distill concise, coherent reasoning segments directly relevant to the model’s final prediction.
Formally, we define a CoT Extraction function $f$ that identifies the minimal yet sufficient reasoning steps required to derive the final answer:
\begin{equation}
    C = f(Q, A, R),
\end{equation}
where $Q$ is the input question, $A$ is the predicted answer, and $R$ is the full reasoning trace. 
The extracted $C$ serves as a condensed and faithful summary of $R$, used as the foundation for all subsequent scoring modules.

To train the extraction function $f$, we construct a synthetic training set via the following pipeline. 
Given a QA pair $(Q, A)$ from a seed question-answering dataset $\mathcal{D}$, a base reasoning model $M_R$ is prompted to generate an initial reasoning trace $R$ and the answer $A$. Another powerful LLM $M_p$ (e.g., GPT-4o~\cite{openai2024gpt4technicalreport}) is then used to summarize $R$ into multiple candidate CoTs ${C_1, \dots, C_k}$, where each candidate is structured with step-wise markers such as \texttt{[STEP]} to highlight the logical progression toward the answer.
To ensure the faithfulness and informativeness of these summaries, we apply a two-stage filtering process. First, we verify that each candidate $C_i$ leads to the correct answer $A$ when used as input alongside $Q$ in $M_p$. Among the valid candidates, we select the most concise one, i.e., the one with the fewest reasoning steps and the shortest token length. Second, to further filter abstraction errors, we assess the semantic alignment between each $C_i$ and the original reasoning trace $R$ using a Natural Language Inference (NLI) classifier, discarding candidates with entailment scores below 0.9, which indicates a potential semantic mismatch.

The resulting $(Q, A, R, C)$ tuples are then used to fine-tune a small LLM $M_s$ to learn the CoT extraction function $f$. The model is trained with a standard language modeling loss over the generated output $C$, conditioning on $(Q, A, R)$ as input. Once trained, the extractor $M_s$ is fixed and used throughout all hallucination detection modules. It is worth noting that the extractor is model-agnostic and task-independent, making it applicable across different LLM settings without requiring re-training. Additional details are provided in Appendix~\ref{apd:extractdetails}.

\subsection{RACE Scoring Framework}
\label{sec:scoring}
Inspired by Equation~\ref{eq1}, we develop our scoring framework, RACE, based on several aspects of consistency. Before computing the hallucination score, the CoT Extractor transforms each $R$ in the main and sampled outputs into $C$, shifting the estimation objective from $H(R,A|Q)$ to $H(C,A|Q)$.

\subsubsection{\texorpdfstring{Reasoning Consistency ($S_{CC}$)}{Reasoning Consistency S\_CC}}

To reduce computational costs of calculating consistency over all pairs of reasoning paths (which has quadratic complexity), we approximate $H(C|Q)$ by measuring the consistency between sampled reasoning paths and the main path. We compare the main CoT ($C_{main}$) with sampled CoTs ($C_i$) on a step-by-step basis, weighting each step by its importance in $C_{main}$:
\begin{equation}
 S_{CC} = \sum_{j} w_j \left(\frac{1}{N}\sum_{i=1}^{N}\delta(C_{\text{main}}^{(j)}, C_i)\right),
 \end{equation}
where $C_{main}^{(j)}$ is the $j$-th step of the main CoT, $C_i$ is one of the sampled CoTs, and $\delta(C_1^{(j)}, C_2)$ denotes the contradiction probability between step $C_1^{(j)}$ and CoT $C_2$, estimated by an NLI classifier, which is generally trained to assess whether the relationship between two sentences is entailment or contradiction (implementation details are shown in the next section).
A higher $S_{CC}$ indicates lower reasoning consistency and thus a greater likelihood of hallucination.

The weight $w_j$ reflects the importance of the $j$-th step of $C_{main}^{(j)}$ to $A_{main}$, aiming to downplay potentially redundant steps after the CoT Extraction process. Specifically, we concatenate the main CoT (with $m$ steps) and main answer as:$$
Q \texttt{ <think> } C_{main}^{(1)}, \dots, C_{main}^{(m)} \texttt{ </think> } A_{main}
.$$

\noindent This sequence is input to an LLM, where the average attention scores from all tokens in $A_{main}$ to those in $C_{main}^{(j)}$ are calculated and normalized to produce weights $w_j$. Detailed calculations are in Appendix~\ref{apd:weightdetails}. We use the CoT Extractor as a proxy to compute attention scores, thus reducing deployment resources and maintaining black-box compatibility with the evaluated LLM.

\subsection{\texorpdfstring{Answer Uncertainty ($S_{AA}$)}{Answer Uncertainty S\_AA}}

We employ SINdex \cite{sindex} to estimate $H(A|Q)$. SINdex refines semantic entropy by adjusting cluster probabilities based on intra-cluster similarity. The SINdex score is calculated as:
\begin{equation}
    S_{AA}  = - \sum_{l=1}^{n} p_l' \log(p_l'),
\end{equation}
where $n$ is the number of clusters from the sampled answers, and $p_l'$ is the adjusted proportion of the $l$-th cluster $C_l$:
\begin{equation}
    p_l' = \frac{p_l \cdot \sum_{x, y \in C_l, x \neq y} \text{sim}(emb(x), emb(y))}{\binom{|C_l|}{2}}.
\end{equation}
Here, $p_l$ is the original proportion of the cluster $C_l$, and $\text{sim}(emb(x), emb(y))$ is the cosine similarity between the embeddings of answers $x$ and $y$ within the cluster. The adjustment factor quantifies semantic coherence within cluster $C_l$. Clusters with high internal dispersion receive a low adjusted proportion $p_l'$. SINdex provides a more accurate measure of uncertainty regarding the answer.

\subsubsection{\texorpdfstring{Reasoning-Answer Alignment ($S_{CA}$)}{Reasoning-Answer Alignment S\_CA}}
To estimate the negative mutual information component, $-I(R, A | Q)$, we assess the alignment between the main reasoning path and the sample answer space generated by the LLM. Specifically, for the main extracted CoT and each sampled answer $A_i$, we use token-level Length-Normalized Predictive Entropy (LNPE)~\cite{malinin2020uncertainty} to quantify how well $C_{main}$ (as context) predicts $A_i$:
\begin{equation}
    S_{CA} = \frac{1}{N} \sum_{i=1}^{N} \bar{H}_{M} ( A_i | Q, C_{main}),
\end{equation}
where $\bar{H}_M(y|x)$ computes the average output token entropy (i.e. LNPE) by feeding $x$ into the LLM $M$ and constraining the output to be $y$.
A higher $S_{CA}$ value suggests poorer alignment between the main reasoning path and the sampled answers, indicating a higher likelihood of hallucination in $O_{main}$. We focus on alignment with $C_{main}$ because our primary goal is to detect hallucinations in the main output, and computing the LNPE score for all sampled reasoning paths and answers leads to quadratic time complexity. The CoT Extractor serves as the  model $M$ for calculating LNPE.

\subsubsection{\texorpdfstring{Reasoning Internal Coherence ($S_{Coh}$)}{Reasoning Internal Coherence S\_Coh}}
We observe that in LRMs, discrepancies can exist between the extracted CoT ($C$) and the original thinking process ($R$). The model's initial reasoning path $R$ may include speculative threads or explorations that do not contribute to the final answer. A high proportion of such speculative content, absent from the concise form $C$, may be associated with an increased rate of hallucinations. Thus, for the LRMs setting, we additionally extract entity sets from the main CoT ($E_{C_{main}}$) and the original main reasoning process ($E_{R_{main}}$). We then compute:
\begin{equation}
S_{Coh} = \left| E_{R_{\text{main}}} \setminus E_{C_{\text{main}}} \right|
    \; \; \big/ \; \left| E_{R_{\text{main}}} \right|,
\end{equation}

\noindent where ``$\setminus$'' denotes the set difference operator. This score quantifies the proportion of entities omitted from the original reasoning in the distilled core CoT. Higher $S_{Coh}$ indicates more speculation and potential hallucination.

\subsubsection{Final Score Aggregation}
Finally, we combine these four metrics into a unified hallucination score for $O_{main}$:
\begin{equation}
    S_{RACE} = S_{AA} + S_{CA} + S_{CC} + S_{Coh},
    \label{equ:race}
\end{equation}
\noindent where the last term is designed specifically for LRMs. This linear combination assigns equal weight to each component, a choice justified by its empirical simplicity and interpretability (Equation~\ref{eq1}). A higher score indicates a greater likelihood that $A_{main}$ contains a hallucination.

\section{Experimental Setup}

\subsection{Datasets and Metrics}
We conduct our evaluation on the validation sets of four widely-used question-answering datasets: TriviaQA~\cite{triviaqa}, SQuAD~\cite{squad}, NQ-Open~\cite{nq}, and HotpotQA~\cite{yang2018hotpotqa}. Following~\citet{su2024unsupervised}, the main output is generated using greedy decoding with a maximum length of 2048 tokens, which represents the model’s default output for detection. For sample-based baselines, we sample 5 outputs using a temperature of 1.0 and top-p sampling with $p=0.95$. For LRMs, main outputs lacking the \texttt{<\textbackslash think>} token are filtered to ensure a final answer is provided. Following~\citet{sindex} and ~\citet{ATTENTIONhallu}, we employ the Area Under the Receiver Operating Characteristic curve (AUROC) as the metric, where a higher AUROC indicates better distinction between hallucinated and non-hallucinated outputs. Following previous works \cite{chen2025un, li-etal-2024-dawn}, we utilize Qwen2.5-32B-Instruct as an LLM-as-Judge to assess whether an LLM's answer contains hallucinations by comparing it to ground truths. Our manual annotations show that LLM-as-Judge can accurately perform hallucination annotation. Comprehensive annotation results, dataset statistics, and all prompts are provided in the Appendix.



\subsection{Model Settings}
\label{sec:models}

RACE accommodates diverse model types and output formats. To evaluate its generality, we consider 3 representative settings reflecting the output styles of different LLM types:

\begin{itemize}
\item \textbf{LRMs:} This is our primary setting, where the model explicitly generates reasoning traces. We evaluate seven LRMs: DeepSeek-R1-Distill-Qwen-7B (DS-7B), DeepSeek-R1-Distill-Llama-8B (DS-8B), DeepSeek-R1-Distill-Qwen-14B (DS-14B), Qwen3-14B (Q3-14B)~\cite{qwen3}, GLM-Z1-9B-0414 (Z1-9B)~\cite{glm2024chatglm}, QwQ-32B\cite{qwq32b}, and DeepSeek-R1~\cite{deepseekai2025deepseekr1incentivizingreasoningcapability}. Inference for DeepSeek-R1 is conducted via its official API.

\item \textbf{CoT Outputs from Standard LLMs:} We evaluate Qwen2.5-14B-Instruct~\cite{qwen2.5} with prompt “Think step by step”, thereby generating the CoT and a final answer. The prompt is detailed in Appendix~\ref{apd:promptcot}.

\item \textbf{Direct Outputs from Standard LLMs:} We evaluate Qwen2.5-14B-Instruct by directly providing the question without any additional prompting.

\end{itemize}

\begin{table*}[t]
\small
\centering
\resizebox{0.95\textwidth}{!}{
\begin{tabular}{ l l c c c  c c c c c c }
\toprule
\textbf{Dataset} & \textbf{Method} & \textbf{DS-7B} & \textbf{DS-8B} & \textbf{DS-14B}  &\textbf{Q3-14B}& \textbf{Z1-9B}& \textbf{QwQ-32B} & \textbf{DS-R1} & \textbf{Q2.5-14B}& \textbf{Q2.5-14B(CoT)}\\
\midrule
\multirow{9}{*}{HotpotQA} & LNPE & 53.60 & 54.01 & 46.79  &61.92& 48.93 & 58.36 & --- & 73.88 & 72.36 \\
 & P(true) & 50.36 & 48.39 & 43.37  &62.60& 73.07 & 66.61 & --- & 73.25 & 67.07 \\
 & SE & 54.98 & 56.48 & 52.95  &56.36& 51.88 & 55.31 & --- & 69.76 & 76.86 \\
 & SEU & 74.66 & 75.37& 76.26&74.75& 73.28 & 76.25 & 73.53 & \underline{75.03} & 74.83 \\
 & SCG & 69.17 & 70.30 & 74.83  &73.43& 68.53 & 69.96 & 67.41 & 57.01 & 71.98 \\
 & SINdex & 74.50 & 74.98 & 76.17  &\underline{75.80}& 74.18& 78.19 & 75.43& 73.26 & \underline{77.19} \\
 & $\text{S}_{\text{RR}}$ & 61.22 & 58.50 & 65.56  &59.65& 58.10 & 66.32 & 53.85 & 60.71 & 54.60 \\
 &  $\text{RACE}_{\text{raw}}$  & \underline{75.48}& \underline{75.58}& \underline{76.82}&75.51& \underline{75.23}& \underline{79.09}& \underline{75.58}& 74.23& 76.71\\
 & RACE & \textbf{77.62} & \textbf{78.56} & \textbf{79.73}  &\textbf{78.41}& \textbf{77.93} & \textbf{81.01} & \textbf{76.71} & \textbf{75.90} & \textbf{79.87} \\
\midrule
\multirow{9}{*}{TriviaQA} & LNPE & 53.70 & 49.93 & 45.91  &69.94& 46.84 & 65.57 & --- & 79.76 & 68.02 \\
 & P(true) & 59.39 & 50.63 & 41.04  &63.86& \underline{83.44} & 78.31 & --- & \underline{86.11} & 85.56 \\
 & SE & 62.33 & 59.66 & 61.10  &65.75& 60.79 & 65.23 & --- & 80.78 & 83.63 \\
 & SEU & 74.42 & 76.73 & 82.47  &86.08& 80.24 & 88.13 & 80.17 & 82.27 & 77.05 \\
 & SCG & 75.42 & 76.76 & \underline{84.83}  &84.37& 73.80 & 78.53 & 51.67 & 67.86 & 86.63 \\
 & SINdex & 77.22& 77.55& 82.47  &87.11& 82.01 & 88.75& \underline{81.56} & 84.30 & \underline{87.33} \\
 & $\text{S}_{\text{RR}}$ & 60.42 & 58.68 & 65.38  &61.48& 61.62 & 70.44 & 52.46 & 73.49 & 69.67 \\
 &  $\text{RACE}_{\text{raw}}$  & \underline{77.36}& \underline{78.27}& 82.72&\underline{87.49}& 82.43& \underline{89.01}& 77.32& 86.06& 86.88\\
 & RACE & \textbf{80.60} & \textbf{81.81} & \textbf{87.03}  &\textbf{89.67}& \textbf{85.54} & \textbf{90.96} & \textbf{83.14} & \textbf{87.02} & \textbf{89.79} \\
\midrule
\multirow{9}{*}{NQ-Open} & LNPE & 55.34 & 54.26 & 52.37  &55.96& 50.40 & 60.04 & --- & 65.60 & 59.22 \\
 & P(true) & 63.45 & 47.76 & 52.98  &63.58& 69.08 & 68.42 & --- & 70.12& 70.01 \\
 & SE & 52.95 & 53.73 & 52.22  &48.10& 45.47 & 50.95 & --- & 62.25 & 67.95 \\
 & SEU & \underline{75.91}& \underline{67.95} & 71.26  &\underline{72.57}& 72.24 & 70.16 & 66.72& 69.92 & 73.75\\
 & SCG & 72.00 & 65.95 & 71.26  &70.88& 66.79 & 65.50 & 61.19 & 62.10 & 71.60 \\
 & SINdex & 72.24 & 65.48 & 70.41  &71.52& 73.19& 72.45 & 66.48 & 66.83 & 71.60 \\
 & $\text{S}_{\text{RR}}$ & 65.98 & 56.63 & 64.26  &62.41& 55.95 & 62.82 & 53.86 & 63.19 & 62.15 \\
 &  $\text{RACE}_{\text{raw}}$  & 75.80& 67.47& \underline{73.14}&68.06& \underline{74.66}& \underline{74.08}& \underline{70.03}& \underline{71.03}& \underline{73.97}\\
 & RACE & \textbf{78.61} & \textbf{72.14} & \textbf{75.80}  &\textbf{75.83}& \textbf{77.81} & \textbf{76.30} & \textbf{73.04} & \textbf{71.29} & \textbf{75.68} \\
\midrule
\multirow{9}{*}{SQuAD} & LNPE & 62.08 & 56.15 & 63.27  &78.73& 63.02 & 78.34 & --- & 80.81 & 76.11 \\
 & P(true) & 47.66 & 52.06 & 54.47  &46.24& 76.95 & 50.75 & --- & 79.16 & 77.56 \\
 & SE & 66.41 & 64.24 & 68.30  &78.30& 74.20 & \textbf{80.33} & --- & \textbf{85.71} & 77.17 \\
 & SEU & \underline{72.06}& \underline{69.74}& 73.72  &73.92& 74.65 & 73.59 & 83.03 & 76.44 & 77.80 \\
 & SCG & 63.16 & 58.41 & 64.63  &71.85& 57.58 & 57.01 & 66.80 & 60.71 & 75.41 \\
 & SINdex & 71.20 & 69.51 & \underline{73.82}&\underline{78.74}& \underline{77.28} & \underline{79.16} & \underline{88.30} & \underline{85.69} & \underline{82.05} \\
 & $\text{S}_{\text{RR}}$ & 55.31 & 48.77 & 49.21  &50.31& 59.13 & 54.07 & 71.23 & 71.61 & 49.75 \\
 &  $\text{RACE}_{\text{raw}}$  & 70.81& 68.91& 72.68&78.34& 74.97& 74.70& 84.71& 84.20& 73.08\\
 & RACE & \textbf{78.27} & \textbf{74.43} & \textbf{79.03}  &\textbf{79.87}& \textbf{79.37} & 77.40 & \textbf{88.95} & 85.02 & \textbf{83.65} \\
\bottomrule
\end{tabular}
}
\caption{Overall AUROC results. DeepSeek-R1 outputs are obtained via the official API, so gray-box methods (LNPE, P(true), SE) cannot detect its hallucinations in this setup. ``Q2.5-14B'' denotes Qwen2.5-14B-Instruct in direct output mode, while ``Q2.5-14B(CoT)'' includes CoT; other models are LRMs. Best and second-best results are in bold and underlined, respectively.}
\label{tab:main_results}

\end{table*}

\subsection{Baselines}
We compare RACE with several zero-resource hallucination detection baselines based on sampling consistency. These include Semantic Entropy (SE)~~\cite{se}, which refines token-level entropy via clustering; SelfCheckGPT-NLI (SCG)~~\cite{manakul2023selfcheckgpt}, which uses an NLI model to identify inconsistencies; Semantic Embedding Uncertainty (SEU)~~\cite{seu}, which assesses consistency through average pairwise embedding similarity; and the Semantic INconsistency Index (SINdex)~~\cite{sindex}, which extends SE by modeling both intra- and inter-class inconsistencies. We also evaluate Length Normalised Predictive Entropy (LNPE)~~\cite{malinin2020uncertainty} and P(true)~~\cite{ptrue}.
For DeepSeek-R1, API access limits evaluation with methods requiring output probabilities (LNPE, P(true), and SE).
These baselines are conducted at the answer level, as semantic clustering or embedding typically fail when applied to lengthy reasoning paths.
Appendix~\ref{apd:baselinedetails} provide details of all baselines.


\subsection{Implementation Details}

For the CoT Extraction module, the seed dataset $\mathcal{D}$ is 2WikiMultihopQA~\cite{2wikiqa}. For dataset construction, we use DeepSeek-Distill-Qwen7B~\cite{deepseekai2025deepseekr1incentivizingreasoningcapability} as the LRM $M_{R}$, Qwen2.5-32B-Instruct as the powerful summarizing model $M_{p}$, and deberta-v3-large-mnli~\cite{manakul2023selfcheckgpt} as the NLI model to filter errors. The extraction model is trained from Llama-3.1-8B-Instruct~\cite{grattafiori2024llama3herdmodels}.\footnote{We have released the model at https://huggingface.co/bebr2/RACE-CoT-Extractor-Llama-8B} We further evaluate the performance of the CoT Extractor in Appendix~\ref{apd:extractorval}.

For $S_{AA}$ (Answer Uncertainty), we align with the SINdex setup, using all-MiniLM-L6-v2~\cite{reimers-gurevych-2019-sentence} as the embedding model.
To adapt lengthy model outputs in our experiments, we adjust the SINdex clustering similarity hyperparameter to 0.9.
For the $S_{CC}$ score (Reasoning Consistency), we use deberta-v3-large-mnli, as employed in SelfCheckGPT, for NLI comparisons. 
For the $S_{Coh}$ score (Reasoning Internal Coherence), we use en\_core\_web\_trf-3.8.0~\cite{spacy2024} to extract entities.

To highlight RACE's contribution, we compare it with two simple methods integrating reasoning process analysis:
\begin{itemize}
    \item \textbf{$\boldsymbol{S_{RR}}$}: A simple adaptation where the CoT component $C$ in $S_{CC}$ is replaced by the model's original reasoning process $R$, segmented by ``\textbackslash n\textbackslash n''.
    \item \textbf{$\text{RACE}_{\text{raw}}$}: A method that  fully aligns with the RACE framework but skips the CoT Extraction module, directly evaluating consistency using the model's original reasoning path (R), equivalent to $S_{AA}$+$S_{RA}$+$S_{RR}+S_{Coh}$.
\end{itemize}

\section{Experimental Results}

We address three key research questions: \textbf{(RQ1)} whether jointly evaluating answer and reasoning provides a useful signal for hallucination detection; \textbf{(RQ2)} whether the structure and noise in initial reasoning paths affect the assessment of reasoning consistency; and \textbf{(RQ3)} how RACE alleviates the delay caused by considering additional reasoning parts.


\subsection{Main Results}

Detailed main results are shown in Table~\ref{tab:main_results}. 
Overall, sampling-based hallucination detection methods outperform probability-based ones, as they incorporate a larger amount of sampling information.
Moreover, for RQ1, the results show that RACE generally outperforms existing sampling-based and probability-based methods across four datasets and seven LRMs in most cases. 
Strong baselines such as SEU and SINdex achieve the best baseline performance on specific models or datasets. However, RACE consistently outperforms them in LRM settings.
Moreover, $\text{RACE}_\text{raw}$ outperforms baselines in many scenarios, achieving the second-best performance, whereas a simple assessment of reasoning consistency ($\text{S}_{\text{RR}}$) yields poor results. This further highlights the effectiveness of the joint evaluation approach.
Furthermore, RACE shows strong generalizability. When applied to Qwen2.5-14B-Instruct generating CoT outputs, RACE significantly exceeds the best baselines. This suggests that RACE’s analysis of LRM’s reasoning component likewise applies to standard instruct models.
Even in the direct output setting, RACE provides competitive or superior performance compared to baselines. 
In this setting, all sampling-based methods utilize the same information. Due to its multi-dimensional scoring and structured extraction of potential reasoning processes, RACE continues to surpass the other baselines.
This indicates that jointly evaluating answer and reasoning consistency is beneficial even when explicit reasoning steps are not prompted.

Regarding RQ2, we compare RACE against $\text{RACE}_\text{raw}$, which aligns with RACE but using the original reasoning paths to evaluate consistency.
The results show that RACE outperforms $\text{RACE}_\text{raw}$ across all models and datasets, indicating that noise within the original reasoning paths compromises consistency assessment. By extracting these paths into more concise, structured CoTs, RACE facilitates the detection of hallucinations associated with the reasoning process.

\subsection{Efficiency Analysis}

\label{cha:efficiency}

\begin{table}[t]
\centering
\small
\setlength{\tabcolsep}{1.5mm}
\resizebox{0.95\linewidth}{!}{
\begin{tabular}{llcc}
\toprule
\multicolumn{2}{l}{}                               & \textbf{Inference Time}    & \textbf{Percentage} \\ \midrule
\multicolumn{2}{l}{\textbf{LLM's Greedy Response}} & 30.91s                     & 100.0\%             \\ \midrule
\multirow{2}{*}{\textbf{SINdex}} & \textbf{LLM's Response} & 43.28s & \multirow{2}{*}{+40.7\%} \\
             & \textbf{SINdex Score}               & +0.02s &                     \\ \midrule
 \multirow{2}{*}{\textbf{$\text{S}_{\text{RR}}$}}& \textbf{LLM's Response} 
& 43.28s &\multirow{2}{*}{+54.3\%}\\
 & \textbf{$\text{S}_{\text{RR}}$ Score}& +4.40s&\\ \midrule
\multirow{3}{*}{\textbf{RACE}}   & \textbf{LLM's Response} & 43.28s & \multirow{3}{*}{+49.4\%} \\
             & \textbf{CoT Extraction}             & +2.10s &                     \\
             & \textbf{RACE Score}& +0.82s &                     \\ \bottomrule
\end{tabular}
}
\caption{Efficiency from generation to detection, showing total time and the percentage of additional time compared to the greedy response. “LLM's Response” includes main and sampled outputs via batch inference. All RACE components are parallelized; $S_{CC}$ is the slowest (0.82s), followed by $S_{CA}$ (0.24s), while the others are negligible.}
\label{tab:efficiency}
\end{table}

To address RQ3, Table~\ref{tab:efficiency} shows the average detection time. We randomly select 100 questions from NQ-Open, using DS-8B with a single 80G GPU.
In sample-based hallucination detection, the primary overhead stems from generating additional responses: producing both main and sampled outputs requires about 40\% more time than the main response alone, while computing the hallucination score adds minimal overhead.
However, reasoning-level hallucination detection introduce more delay than answer-level methods, as reasoning paths are generally lengthy and complex. RACE simplifies the reasoning path using the CoT Extraction module, achieving significantly better performance than the naive baseline ($\text{S}_{\text{RR}}$) with fewer additional delays.
Additionally, RACE involves deploying a 8B CoT Extractor. Compared with larger state-of-the-art reasoning LLMs (for example, DeepSeek-R1 at 617B), this deployment overhead is acceptable. For applications demanding high reliability, the trade-off between overhead and accuracy is often needed.

\begin{table}[t]
\centering
\small
\resizebox{\linewidth}{!}{
\setlength{\tabcolsep}{1mm}
\begin{tabular}{cccccc}
\toprule
\multicolumn{1}{l}{}              & \textbf{DS-7B} & \textbf{DS-8B} & \textbf{DS-14B} & \textbf{Z1-9B} & \textbf{QwQ-32B} \\ \midrule
\multicolumn{1}{l}{\textbf{RACE}} & \textbf{77.62} & \textbf{78.56} & {\underline{79.73}}     & \textbf{77.93} & \textbf{81.01}   \\
\textbf{w/ avg. $S_{CC}$} & 77.09 & 78.02 & 79.18          & {\underline{77.57}} & 80.81 \\ \midrule
\textbf{w/o $S_{Coh}$}            & {\underline{77.48}}    & {\underline{78.15}}    & 79.46           & 77.51          & {\underline{80.86}}      \\
\textbf{w/o $S_{AA}$}     & 75.43 & 77.79 & 79.50          & 77.45       & 80.02 \\
\textbf{w/o $S_{CA}$}     & 77.26 & 77.82 & 79.50          & 77.48       & 80.49 \\
\textbf{w/o $S_{CC}$}     & 75.08 & 76.08 & 76.40          & 75.24       & 78.97 \\ \midrule
\textbf{only $S_{Coh}$}   & 51.07 & 56.95 & 56.08          & 55.98       & 57.39 \\
\textbf{only $S_{AA}$}    & 74.50 & 74.98 & 76.17          & 74.18       & 78.19 \\
\textbf{only $S_{CA}$}    & 68.83 & 68.76 & 67.33          & 68.35       & 69.43 \\
\textbf{only $S_{CC}$}    & 74.98 & 75.49 & \textbf{79.81} & 76.76       & 79.46 \\ \bottomrule
\end{tabular}
}
\caption{The ablation study of each component on HotpotQA. ``\textit{w/ avg. $S_{CC}$}'' means when calculating $S_{CC}$, the importance weighting for each step is not applied.}
\label{tab:ablation}
\end{table}

\subsection{Ablation Study}

\textbf{Ablation of each component:} Table~\ref{tab:ablation} reveals the contribution of each component in RACE. Removing any single component generally leads to a decrease in performance. Using a single component results in lower performance, confirming that all components contribute complementary signals for effective detection. Note that using only the Reasoning Consistency score (only $S_{CC}$) achieves the best performance among all single components, highlighting the importance of detecting consistency among reasoning paths. 

\noindent \textbf{Weight optimization of each component:} In our main experiment, we assign equal weights to each RACE component for interpretability (Equation~\ref{eq1}) and to avoid fine-tuning that may lead to unfair comparisons. Table~\ref{tab:optim} shows further gains from weight optimization. We use the first 20\% of each dataset for training and the rest for testing. On the training set, we perform a grid search over [0, 1] with 0.05 increments, normalize the best-performing weights, and evaluate on the test set. Results (the column of RACE$^+$) show that optimized weights can further improve RACE, though the best configurations vary across models and datasets, which suggest that RACE has strong potential in real-world applications. The results also demonstrate that despite their varying importance, all four components are indeed effective.

\textbf{In Appendix~\ref{apd:abl}, we provide more ablation study and further analysis}, including the impact of clustering hyperparameters, CoT Extraction module training settings, and the number of samples. We also investigate the the necessity of training the CoT Extraction module and the plug-and-play capabilities of RACE, and provide the case study.

\begin{table}[]
\centering

\resizebox{\linewidth}{!}{
\small
\setlength{\tabcolsep}{1mm}
\begin{tabular}{lccc}
\toprule
                     & \textbf{RACE} & \textbf{RACE$^+$} & \textbf{Optimal Weights}  \\ \midrule
\textbf{Q3-14B(NQ)}   & 76.55         & \textbf{79.71}   & 0.09 / 0.49 / 0.37 / 0.06 \\ \midrule
\textbf{Q3-14B(HotpotQA)} & 78.33         & \textbf{78.76}   & 0.13 / 0.35 / 0.29 / 0.23 \\ \midrule
\textbf{DS-8B(NQ)}    & 72.30         & \textbf{76.82}   & 0.03 / 0.34 / 0.34 / 0.28 \\ \midrule
\textbf{DS-8B(HotpotQA)}  & 78.46         & \textbf{79.73}   & 0.10 / 0.35 / 0.33 / 0.22 \\ \bottomrule
\end{tabular}
}
\caption{AUROC values (first two columns) and normalized optimal weights (last column) from optimizing RACE component coefficients (RACE$^+$), ordered as in Equation~\ref{equ:race}.}
\label{tab:optim}
\end{table}

\section{Conclusion and Future Work}

In this work, we introduce RACE, a novel black-box hallucination detection framework. By jointly assessing the consistency of the reasoning process and the final answer, RACE provides fine-grained detection of hallucinations. Experimental results demonstrate that RACE significantly outperforms existing methods across multiple datasets and LLMs, highlighting its robustness and generalizability for hallucination detection.

While this work focuses on hallucination detection, mitigating hallucinations remains an equally important research area~\cite{su2024mitigating}. 
In future work, we plan to explore hallucination mitigation strategies that build on the RACE framework. 
For instance, our method can naturally be adapted to select the most reliable reasoning path from multiple generations, thereby reducing the likelihood of hallucinations at inference time. 
Moreover, Retrieval-Augmented Generation (RAG) is also a promising research area for mitigating hallucinations~\cite{lewis2020retrieval,su2024dragin,tu2025rbft,dong2025decoupling,su2025parametric,su2025judge}. 
It operates by first retrieving relevant documents using either standard retrievers~\cite{zhai2008statistical,fang2024scaling,su2023thuir2,su2023wikiformer,ma2023caseencoder,su2023caseformer} or more advanced retrieval systems~\cite{salemi2024towards,li2023towards,ye2024relevance,chen2022web,su2024stard}, and then appending them to the LLM’s input to provide external evidence for generation. 
Applying RACE within RAG-based systems allows hallucinations to be mitigated through both retrieval-based grounding and reasoning consistency assessment, representing a promising direction for future research.




\bibliography{custom}

\newpage
\appendix

\section{Details of RACE Framework}
\subsection{Details of the CoT Extractor}
\label{apd:extractdetails}
We first employ the DeepSeek-Distill-Qwen7B model on the 2WikiMultihopQA validation set, comprising 12,576 questions, using greedy decoding to derive the reasoning process and answers. The model prompt is as follows:

\begin{tcolorbox}[colback=lightgray!20,colframe=darkgray!80,title=Prompts for training data generation]
Answer the user's question. You should reason step by step first. Each reasoning step should start with [STEP]. Finally, output [ANSWER] and provide the final answer. The reasoning steps should be complete, meaning they should include the entire thought process from the question to the final answer, but they should also be concise, meaning there should be no redundant reasoning steps.
\\ \hspace*{\fill} \\
Example:
Do both films The Falcon (Film) and Valentin The Good have the directors from the same country?
\\ \hspace*{\fill} \\

[STEP] The director of The Falcon (film) is Vatroslav Mimica.

[STEP] Vatroslav Mimica is from Croatian.

[STEP] Valentin The Good's director is Martin Frič.

[STEP] Martin Frič is from Czech.

[STEP] So they are from the differenct country.

[ANSWER] No.
\end{tcolorbox}

Then, for each question, along with the reasoning process and answer generated by the model, we use Qwen2.5-32B-Instruct to extract the most critical elements of the reasoning. We generate six summaries accordingly. We use the following system prompt, with the "user" input section formatted in the same way as the provided example.

\begin{tcolorbox}[colback=lightgray!20,colframe=darkgray!80,title=System prompts for CoT Extraction]
You are responsible for extracting the main steps of the CoT (Chain of Thought). For the user's input question, thought process, and final answer, extract the steps in the thought process that lead to the final answer, ignoring irrelevant exploration or backtracking steps, merging the same step, and separating adjacent reasoning steps with [STEP].
Your output should only contain the extracted CoT and must be faithful to the user's input, even if the thought process contains errors. Minimize the number of inference steps and keep each step concise.
\\ \hspace*{\fill} \\
Example:
\\ \hspace*{\fill} \\
\#\# Question

Were Scott Derrickson and Ed Wood of the same nationality?"
\\ \hspace*{\fill} \\
\#\# Thought

Okay, so the user is asking if Scott Derrickson and Ed Wood are from the same country. I remember Scott Derrickson is an American director, known for horror movies like "Sinister.\ Ed Wood, on the other hand, I think is also American, but I'm not 100\% sure. Maybe I should double-check that. I recall that Ed Wood was a filmmaker in the 50s and 60s, known for low-budget movies, so he's probably from the US. Yeah, I think both are American. So, the answer is yes, they're both from the same nationality, which is American.
\\ \hspace*{\fill} \\
\#\# Final Answer

Yes, both Scott Derrickson and Ed Wood were American nationals.
\\ \hspace*{\fill} \\
\#\# Output

[STEP] Scott Derrickson is an American director, known for horror movies like "Sinister."

[STEP] Ed Wood was a filmmaker in the 50s and 60s, known for low-budget movies, and is likely from the US.

[STEP] Both Scott Derrickson and Ed Wood are American.

[STEP] So the answer is: yes, Scott Derrickson and Ed Wood are of the same nationality.
\end{tcolorbox}

We then conduct three stages of filtering on the generated data. For each question \(Q\), the LRM’s reasoning process \(R\), its final answer \(A_{R}\), and the distilled set of \(k\) CoTs \((C_1, \ldots, C_k)\) where \(k = 6\):
\begin{enumerate}
    \item We prompt Qwen2.5-32B-Instruct to produce an answer \(A_i\) based on \(Q\) and \(C_i\) for \(i = 1, 2, \ldots, k\). We compare \(A_i\) with \(A_R\) to check for equivalence and keep only those \(C_i\) whose answers match \(A_R\).  
    \item From all the retained \(C_i\), we take the half with the fewest words and then select the one with the fewest steps, calling it \(C_{concise}\).  
    \item We perform an NLI computation (using deberta-v3-large-mnli model) for each step of \(C_{concise}\) against the original LRM reasoning process \(R\). Only when every step’s entailment score is above 0.9 does \(C_{concise}\) enter the training dataset.
\end{enumerate}

Through this procedure, we obtain \((Q, R, A, C_{concise})\) tuples, ensuring that \(C_{concise}\) remains faithful to both the LRM’s reasoning process and final answer. We then use this dataset to train Llama3.1-8B-Instruct, adopting the same prompt format used by Qwen2.5-32B-Instruct for producing the distilled CoT. We calculate cross-entropy loss only on \(C_{concise}\). In the main experiment, the CoT Extractor dataset consists of 2000 examples, and we train for two epochs. The impact of training settings is detailed in Section~\ref{apd:absextractor}.

\subsection{Details of weight calculations in $S_{CC}$}
\label{apd:weightdetails}

The weights $w_j$ are designed to quantify the importance of each step $C_{main}^{(j)}$ in the $C_{main}$ to the main answer $A_{main}$. This mechanism helps in reducing the influence of potentially redundant steps that may arise during the CoT extraction process. We start with inputting the following sequence to the LLM (in our experiments, it is the CoT Extractor, as a proxy model), concatenating the question ($Q$), the extracted CoT steps ($C_{main}^{(1)}, \dots, C_{main}^{(k)}$), and the main answer ($A_{main}$) as follows:
$$ Q \texttt{ <think> } C_{main}^{(1)}, \dots, C_{main}^{(k)} \texttt{ </think> } A_{main} $$

Let $A_{main}$ be represented by its sequence of $M$ tokens, $(a_1, a_2, \dots, a_M)$. Similarly, let the $j$-th CoT step, $C_{main}^{(j)}$, be represented by its sequence of $L_j$ tokens, $(c_{j,1}, c_{j,2}, \dots, c_{j,L_j})$.
The core of the weight calculation lies in leveraging the attention mechanism of the LLM. For each step $C_{main}^{(j)}$, we compute an aggregate attention score, denoted as $S_j$. This score represents the average attention from all tokens in the main answer $A_{main}$ to all tokens in the step $C_{main}^{(j)}$. The term $\text{Attn}(a_p, c_{j,q})$ refers to the attention score from token $a_p$ (in $A_{main}$) to token $c_{j,q}$ (in $C_{main}^{(j)}$). This individual attention score $\text{Attn}(a_p, c_{j,q})$ is itself an average of the attention values across all layers and attention heads of the LLM for that token pair, given the constructed input sequence.

The aggregate attention score $S_j$ for the $j$-th step is formulated as:
$$ S_j = \frac{1}{M \cdot L_j} \sum_{p=1}^{M} \sum_{q=1}^{L_j} \text{Attn}(a_p, c_{j,q}) $$
This formula effectively calculates the mean attention interaction between the entirety of the main answer and the entirety of the $j$-th step.

After computing these raw scores $S_j$ for all $k$ steps, they are normalized to produce the final weights $w_j$. The weight $w_j$ for the $j$-th step is calculated as:
$$ w_j = \frac{S_j}{\sum_{i=1}^{k} S_i} $$
These resulting weights $w_j$ are then used to modulate the contribution of each CoT step, fulfilling the objective of emphasizing more relevant steps and downplaying less critical or redundant ones.

\section{Details of Experimental Setup}

\subsection{Details of Evaluation}
\label{apd:datasetdetails}
\subsubsection{Details of Evaluation Datasets}

We conduct our evaluation on the validation sets of four widely used question-answering datasets:

\begin{itemize}
    \item \textbf{TriviaQA}~\cite{triviaqa}: A challenging dataset featuring diverse trivia questions requiring broad world knowledge.
    \item \textbf{SQuAD}~\cite{squad}: A standard reading comprehension benchmark based on Wikipedia articles. We utilize version 2.0 and exclude any questions labeled unanswerable.
    \item \textbf{NQ-Open}~\cite{nq, lee-etal-2019-latent}: An open-domain QA dataset based on search queries, focusing on realistic needs.
    \item \textbf{HotpotQA}~\cite{yang2018hotpotqa}: A multi-hop QA dataset requiring reasoning to reach the answer.
\end{itemize}

TriviaQA, NQ-Open, and HotpotQA employ a closed-book question-answering format, with dataset sizes of 7993, 1800, and 7405, respectively. SQuAD is in a reading comprehension format. We remove questions labeled as unanswerable, leaving 5928 questions.

\subsubsection{Details of prompts for LLM-as-Judge}
Following \citet{chen2025un}, we employ Qwen2.5-32B-Instruct to assess whether outputs from LLMs contain hallucinations. Specifically, we use a multiple-choice prompt, selecting the option with the highest output probability to represent the model's decision. If option A is chosen, it is marked as free of hallucinations; any other choice indicates the presence of hallucinations. The prompt is as follows:

\begin{tcolorbox}[colback=lightgray!20,colframe=darkgray!80,title=Prompts for labeling hallucinations]
I will provide a question, ground truth, and an answer. You need to determine whether the answer is correct. Choose the most appropriate option from the following:
\\ \hspace*{\fill} \\
A. Correct: The answer is semantically equivalent to the ground truth.

B. Incorrect: The answer addresses the question but is not semantically equivalent to any of the ground truths.

C. Irrelevant: The answer is unrelated to the question or does not provide a valid response.
\\ \hspace*{\fill} \\
Question: \{\textit{A question}\}

Ground Truth: \{\textit{A golden answer}\}

Answer: \{\textit{An answer generated by LLMs}\}
\\ \hspace*{\fill} \\
Your choice:
\end{tcolorbox}

\subsubsection{Details of prompts for generation}
\label{apd:promptcot}
We evaluate three generation setups. For \textbf{``LRM''} and \textbf{``Direct Outputs from Standard LLMs''}, we simply supply the question to the LLMs without any extra prompts. For \textbf{``CoT Outputs from Standard LLMs''}, we use the following prompt, prompting the model to generate the chain of thought before providing the final answer.

\begin{tcolorbox}[colback=lightgray!20,colframe=darkgray!80,title=Prompts for CoT Outputs from Standard LLMs]
Question: \{\textit{A question}\}
\\ \hspace*{\fill} \\
- Think step by step using the format: "Thought: ..."
- Then conclude with the answer using the format: "Answer: ..."
\end{tcolorbox}

\subsection{Details of Baselines}
\label{apd:baselinedetails}
We adopt the original models proposed in the respective papers for sample consistency-based methods. Specifically, Semantic Entropy uses the \texttt{deberta-large-mnli} model~\cite{he2021debertav3}, SelfCheckGPT-NLI uses the \texttt{deberta-v3-large-mnli} model, and both SEU and SINdex employ the \texttt{all-MiniLM-L6-v2} model for embeddings. Moreover, SINdex shares the same hyperparameter setting as RACE, set to 0.9. The impact of this hyperparameter on SINdex and RACE is presented in Section~\ref{apd:abshyper}. For P(true), we use the zero-shot prompt provided in the original paper.

\section{CoT Extractor Validation}
\label{apd:extractorval}

\begin{table*}[t!]
\centering
\resizebox{\textwidth}{!}{
\begin{tabularx}{\textwidth}{X}
\toprule
\textbf{Question:} What is the birthday of the singer whose songs are remixed in Queen of Clubs Trilogy: Onyx Edition? \\ \midrule
\textbf{Reasoning (Simplified):}\\
Okay, so I need to figure out the birthday of the singer whose songs are remixed in the Queen of Clubs Trilogy: Onyx Edition. Hmm... Let me think about this step by step.
\\ \hspace*{\fill} \\
First, I know that the *Queen of Clubs Trilogy: Onyx Edition* is a music album, likely a remix collection... So, I need to find out which singer's tracks are remixed...
\\ \hspace*{\fill} \\
I'm not very familiar with the specific artists... Names like Madonna, Michael Jackson, or maybe... Britney Spears or Justin Timberlake come to mind... But I'm not sure...
\\ \hspace*{\fill} \\
Wait, perhaps it's a more niche artist or someone from the electronic or dance scene... Maybe someone like The Killers...?
\\ \hspace*{\fill} \\
Alternatively, maybe it's a singer known for having their songs remixed a lot... there's the "Remix" series... But I'm not sure...
\\ \hspace*{\fill} \\
I should also consider the title: *Queen of Clubs Trilogy: Onyx Edition*... Maybe it refers to a stage name...
\\ \hspace*{\fill} \\
Wait, I think I might have heard that the album features songs by a particular artist... Maybe it's someone like Dua Lipa?... Or Taylor Swift?...
\\ \hspace*{\fill} \\
I'm also considering it might be a specific remix album... but I don't think that's the one...
\\ \hspace*{\fill} \\
Wait, perhaps it's a specific artist known for their remixes... But again, I'm not sure...
\\ \hspace*{\fill} \\
I think the *Queen of Clubs Trilogy* might feature remixes by a singer born in the late '70s... or maybe early '90s...
\\ \hspace*{\fill} \\
Wait, I think I've heard the singer's birthday is in September... Maybe September 5th...
\\ \hspace*{\fill} \\
I'm not entirely confident... So, I'll go with that as my best guess. \\

\midrule
\textbf{Answer:} The singer whose songs are remixed in the Queen of Clubs Trilogy: Onyx Edition is believed to have a birthday in September, possibly around the 5th. \\
  \midrule
\textbf{CoT by CoT Extractor: }   \\               The Queen of Clubs Trilogy: Onyx Edition is a music album featuring remixed songs. \textit{(Weight: 0.1594)}\\
The user asks for the birthday of the singer whose songs are remixed in the album. \textit{(Weight: 0.2161)}\\
The album is associated with a singer whose birthday is in September. \textit{(Weight: 0.2298)}\\
The specific birthday is believed to be around the 5th of September. \textit{(Weight: 0.3946)}\\
\bottomrule
\end{tabularx}
}
\caption{Results of CoT extraction from DeepSeek-R1-Distill-Llama-8B and HotpotQA. Weights mean the computed importance weights of each CoT step for the $S_{CC}$ component.}
\label{tab:casecot}
\end{table*}

To further evaluate the trained CoT Extraction module, we carried out manual annotations on 250 results randomly sampled from DeepSeek-R1-Distill-Qwen-7B and DeepSeek-R1-Distill-Llama-8B in the NQ-Open and HotpotQA datasets. We began by assessing the faithfulness of the extracted CoTs, defining faithfulness as the absence of any alterations or additions to the original reasoning path and answer, while preserving the integrity of the core reasoning (excluding redundancy considerations). The annotations revealed that only one of 250 samples was deemed unfaithful, corresponding to a faithfulness rate of 99.6\%. This single instance resulted solely from a misexpression in a redundant step, underscoring the CoT Extractor’s robustness.

Furthermore, to examine potential redundancies within CoTs, we conducted a case study presented in Table~\ref{tab:casecot}. Although the first two steps in this example are evidently redundant, the computation of $S_{CC}$ assigns varying weights to each step, granting higher importance to the latter two core reasoning steps. This weighted approach, combined with additional strategies, effectively mitigates the impact of redundant reasoning in CoTs, demonstrating the overall robustness of the RACE method.

\section{Further Ablation Studies}
\label{apd:abl}
\begin{figure}
    \centering
    \includegraphics[width=\linewidth]{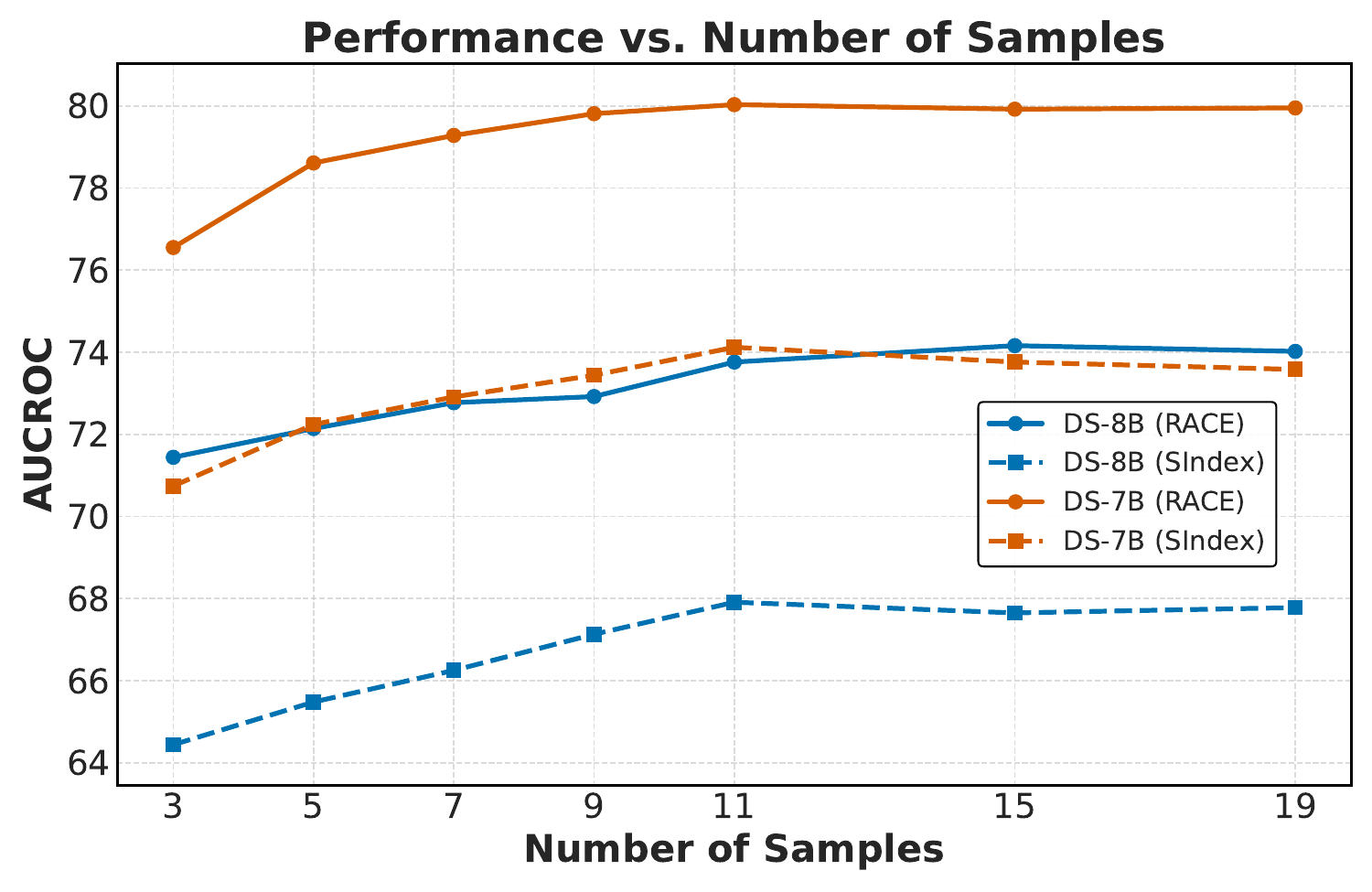}
    \caption{AUROC performance of RACE and SINdex across different numbers of generations on NQ-Open.}
    \label{fig:samples}
\end{figure}

\subsection{Impact of Number of Samples}
\label{apd:numbers}
Figure~\ref{fig:samples} illustrates the relationship between the number of generated samples and detection performance. We compare RACE with a baseline that utilizes only $S_{AA}$ (i.e., SINdex). Both detection methods exhibit improved performance as the number of generated samples increases, reaching their peak at approximately 11 samples; however, RACE consistently performs better than SINdex.

\subsection{Impact of Training or Simple Prompting on CoT Extractor}

Table~\ref{tab:cotextractor} investigates the performance differences between a trained CoT Extractor and its base LLM when using the same prompt. The trained CoT Extractor offers a slight improvement in hallucination detection, and more notably, it substantially shortens the extracted CoT. This reduction leads to fewer output tokens (and thus shorter output time) and lowers the computational overhead for subsequent hallucination detection.

\begin{table}[!h]
\centering 
\resizebox{0.99\linewidth}{!}{
\label{tab:transposed_results} 
\begin{tabular}{lcccc}
\toprule
\textbf{CoT Extractor} & \multicolumn{2}{c}{\textbf{Trained Extractor}} & \multicolumn{2}{c}{\textbf{Llama3.1-8B}} \\ \midrule
\textbf{Model}         & \textbf{AUROC}     & \textbf{Avg. Tokens}     & \textbf{AUROC}  & \textbf{Avg. Tokens}  \\ \midrule
\textbf{DS-7B}   & 78.61 & 84.57  & 78.32 & 342.91 \\
\textbf{DS-8B}   & 72.14 & 85.63  & 71.89 & 340.90 \\
\textbf{DS-14B}  & 75.80 & 85.35  & 74.79 & 333.72 \\
\textbf{Z1-9B}   & 77.81 & 94.02  & 77.31 & 396.45 \\
\textbf{QwQ-32B} & 76.30 & 100.49 & 70.62 & 418.22 \\ \bottomrule
\end{tabular}
}
\caption{Comparison between the trained CoT Extractor and its original model using the same prompt on NQ. \textit{``Avg. Tokens''} indicates the average output length per extraction.}
\label{tab:cotextractor}
\end{table}

\subsection{Impact of Training Dataset Size of the CoT Extractor}
\label{apd:absextractor}
\begin{table}[h]
    \centering
    \resizebox{\linewidth}{!}{
    \begin{tabular}{lccccc}
\toprule
\textbf{Dataset Size} & \textbf{100} & \textbf{500} & \textbf{1000} & \textbf{2000} & \textbf{4000} \\ \midrule
\textbf{AUROC}       & 79.13        & 79.46        & 79.84         & 79.73         & 79.62         \\ \bottomrule
\end{tabular}
    }
     \caption{The relationship between the CoT Extractor’s training dataset size and the final performance on HotpotQA, using DS-14B.}
    \label{tab:dts}
\end{table}

We compare the relationship between the training dataset size of the CoT Extractor and the final RACE detection performance, as shown in Table~\ref{tab:dts}. We find that performance increases substantially before the dataset size reaches 1000 and peaks around 1000. However, the dataset size exerts only a modest influence on the overall performance.
\subsection{Impact of Similarity Threshold for Clustering in SINdex ($S_{AA}$)}
\label{apd:abshyper}
\begin{table}[h]
    \centering
    \resizebox{\linewidth}{!}{
    \begin{tabular}{lcccccc}
\toprule
\textbf{Threshold} & \textbf{0.95} & \textbf{0.9} & \textbf{0.85} & \textbf{0.8} & \textbf{0.75} & \textbf{0.7} \\ \midrule
\textbf{SINdex}    & 72.79         & 74.98        & 74.38         & 74.37        & 74.49         & 74.61        \\
\textbf{RACE}      & 78.16         & 78.56        & 78.16         & 78.38        & 78.62         & 78.60        \\ \bottomrule
\end{tabular}
}
\caption{Impact of the Similarity Threshold hyperparameter on performance, using the HotpotQA dataset and the DS-8B model.}
    \label{tab:threshold}
\end{table}

Table~\ref{tab:threshold} illustrates how the hyperparameter changes affect RACE and SINdex. This hyperparameter controls the similarity threshold that groups two outputs into the same cluster. Because RACE includes extra components to detect reasoning hallucinations, it is less sensitive to this hyperparameter than SINdex: when the threshold is set as high as 0.95, SINdex experiences a pronounced decline in performance, whereas RACE displays only a minor decrease, highlighting its robustness. Both methods stabilize when the threshold is less than or equal to 0.9.

\subsection{Plug-and-play Analysis}

\begin{table}[h!]
\centering
\resizebox{\linewidth}{!}{
\begin{tabular}{lcccccc}
\toprule
\textbf{$S_{AA}$}                & \textbf{$S_{CC}+S_{CA}+S_{Coh}$} & \textbf{DS-7B} & \textbf{DS-8B} & \textbf{DS-14B} & \textbf{Z1-9B} & \textbf{QwQ-32B} \\ \midrule
\multirow{2}{*}{\textbf{SINdex}} & \cmark                       & \textbf{77.62} & \textbf{78.56} & \textbf{79.73}  & \textbf{77.93} & \textbf{81.01}   \\
                              & \xmark & 74.50          & 74.98          & 76.17          & 74.18          & 78.19          \\ \midrule
\multirow{2}{*}{\textbf{SEU}} & \cmark & \textbf{78.08} & \textbf{79.60} & \textbf{80.70} & \textbf{79.15} & \textbf{81.03} \\
                              & \xmark & 74.66          & 75.37          & 76.26          & 73.28          & 73.28          \\ \midrule
\multirow{2}{*}{\textbf{SCG}} & \cmark & \textbf{76.10} & \textbf{77.89} & \textbf{80.46} & \textbf{77.67} & 79.51 \\
                              & \xmark & 69.17          & 70.30          & 74.83          & 68.53          & 69.96          \\ \midrule
\xmark                 & \cmark          & 75.43          & 77.79          & 79.50          & 77.45          & 80.02          \\ \bottomrule
\end{tabular}
}
\caption{The plug-and-play analysis of RACE on HotpotQA. Adding RACE's reasoning-focused components to existing sampling-based methods can enhance the overall performance. We bold the combined result if it surpasses both pre-combination outcomes.}
\label{tab:plugandplay}
\end{table}

Table~\ref{tab:plugandplay} illustrates RACE's effectiveness as an enhancement module when the scores ($S_{CC}+S_{CA}+S_{Coh}$) are integrated into existing sampling-based methods. These results indicate that the reasoning consistency signals captured by RACE can effectively augment existing hallucination detection frameworks.
Additionally, Answer Uncertainty ($S_{AA}$) is also crucial, as methods that solely rely on reasoning components typically underperform. 
This is likely because shorter answers allow for more accurate uncertainty estimation.

\subsection{Further Analysis of RACE$^+$}

In Table~\ref{tab:optim}, we compare RACE$^+$ (RACE with optimized component weights) to its equal-weighted counterpart. Here is further analysis of the results:

First, coefficient magnitude does not directly equate to component importance. $S_{AA}$ relies on the SINdex score, an entropy-based metric that can exceed 1, whereas $S_{CC}$ and $S_{Coh}$ are strictly within [0, 1]. Although $S_{CA}$ is also entropy-based, our DS-8B and NQ analyses show it exceeds 1 only 6.5\% of the time, versus 64.0\% for $S_{AA}$. This justifies $S_{AA}$’s lower optimal weight given its broader range.

Second, the optimal weights confirm that $S_{CA}$ and $S_{CC}$, which capture reasoning-related information, are more influential and thus receive higher weights. This likely reflects the interdependence of terms in the decomposition of $H(R, A \mid Q)$: $R$ often incorporates information from $A$, making reasoning detection particularly critical, as shown by the “only $S_{CC}$” setting’s strong performance in Table~\ref{tab:ablation}. Nonetheless, $A$ can also contain information absent from $R$, and both Table~\ref{tab:ablation} and Table 4~\ref{tab:optim} underscore that $S_{AA}$ remains essential.

Finally, our main experiments employ equal weights to prevent any unfair tuning and to maintain alignment with Equation 1. Yet Table 4~\ref{tab:optim} highlights the significant gains of RACE$^+$, maybe caused by varying reasoning behaviors across models and differences in downstream data distributions. We therefore strongly recommend calibrating component weights on small-scale datasets to further boost RACE’s performance. Crucially, even the equal-weighted RACE already surpasses existing baselines, demonstrating RACE’s robustness.

\section{Case Study}

\begin{table*}[t]
\resizebox{\textwidth}{!}{
\begin{tabularx}{\textwidth}{lYX}
\toprule
\multicolumn{3}{l}{\textbf{Question:} Which tennis player was born first, Kimberly Po or Leoš Friedl? \ \ \ \ \textbf{Golden Answer:} Leoš Friedl} \\ \midrule
\textbf{Output}                  & \multicolumn{1}{Y}{\textbf{CoT by CoT Extractor}}          & \multicolumn{1}{X}{\textbf{Answer}}         \\ \midrule
\textbf{Main}          &    
Kimberly Po was born on July 15, 1979, in the United States.

Leoš Friedl was born on April 15, 1981, in the Czech Republic.

Comparing their birth years, Kimberly Po was born earlier than Leoš Friedl.
&   Kimberly Po was born first. ...                   \\ \midrule
\textbf{Sample 1}               &   
Kimberly Po was born on November 16, 1979.

Leoš Friedl was born on July 21, 1980.

Comparing their birth years, Kimberly Po was born first.
&                 Kimberly Po was born first. ...                            \\ \midrule
\textbf{Sample 2}               &    
Kimberly Po was born on January 10, 1978.

Leoš Friedl was born on September 11, 1978.

Since January 10, 1978, is earlier than September 11, 1978, Kimberly Po was born first.
&        Kimberly Po was born first. ...                                     \\ \midrule
\textbf{Sample 3}               &
Kimberly Po was born on March 25, 1979.

Leoš Friedl was born on April 15, 1983.

Comparing their birth years, Kimberly Po was born earlier than Leoš Friedl.&     Kimberly Po was born first. ...                                        \\ \midrule
\textbf{Sample 4}               &
Kimberly Po was born on April 27, 1976.

Leoš Friedl was born on October 13, 1979.

Comparing their birth dates, Kimberly Po was born before Leoš Friedl.
&           Kimberly Po was born first. ...                                  \\ \midrule
\textbf{Sample 5}               &
Kimberly Po was born on June 25, 1979, in New York, USA.

Leoš Friedl was born on May 23, 1978, in Plzeň, Czechoslovakia (now the Czech Republic).

Comparing their birth years, Leoš Friedl was born before Kimberly Po.&          Leoš Friedl was born first. ...               \\ \midrule
\multicolumn{3}{l}{\textbf{SINdex:} 0.0383 (Normalized: 0.0936)\ \ \ \ \textbf{RACE:} 2.1571  (Normalized: 0.3685)}

\\ \bottomrule
\end{tabularx}
}
\caption{A hallucination example drawn from Qwen3-14B and the HotpotQA dataset. Nonessential parts of the answers have been omitted, and the original reasoning has been condensed into the summarized CoT for clarity. The last line, ``Normalized'', represents the percentile rank of each detection score across all instances in this dataset and model, where 1 corresponds to the maximum score, signifying the greatest likelihood of hallucination.}
\label{tab:case}
\end{table*}

Figure~\ref{fig:main} already presents a real-world example. Here, we introduce another challenging case, shown in Table~\ref{tab:case}. Although the main output contains hallucinations, an answer-level method like SINdex assigns a low hallucination score because four out of five samples match the main answer. This score ranks in the bottom 10\% of the dataset, implying no hallucination. In contrast, RACE analyzes the full reasoning chain and detects inconsistent birthdays across every reasoning path, raising the normalized hallucination score from 0.0936 to 0.3685. This case highlights the advantage of RACE’s combined reasoning and answer-level evaluation over pure answer-level detection.

\newpage
\section{Licensing}
Qwen2.5-14B-Instruct, Qwen2.5-32B-Instruct, and QwQ-32B are released under the Apache License 2.0. Llama-3.1-8B-Instruct is released under the META LLAMA 3 COMMUNITY LICENSE. GLM-Z1-9B-0414 and all DeepSeek-series models we used are released under the MIT License.

The datasets HotpotQA and SQuAD are released under the CC BY-SA 4.0 License. TriviaQA, NQ-Open, and 2WikiMultihopQA are released under the Apache License 2.0. 

This paper's research objective is academic exploration, which aligns with the terms of these licenses.

\end{document}